\documentclass{article}
\usepackage{ijcai20}

\usepackage{times}
\usepackage{soul}
\usepackage{url}
\usepackage[hidelinks]{hyperref}
\usepackage[utf8]{inputenc}
\usepackage[small]{caption}
\usepackage{graphicx}
\usepackage{subfigure}
\usepackage{amsmath}
\usepackage{amsfonts}
\usepackage{amsthm}
\usepackage{booktabs}
\usepackage{algorithm}
\usepackage{algorithmic}
\usepackage{multirow}
\usepackage{xspace}
\usepackage[english]{babel}
\newcommand{\TriNN}{\textsc{Deal}\xspace}

\title{ Inductive Link Prediction for Nodes Having Only Attribute Information } %on Attributed Graphs}

\author{
Yu Hao$^1$\and
Xin Cao$^1$\and
Yixiang Fang$^1$\footnote{Corresponding author}\and
Xike Xie$^2$ \And
Sibo Wang$^3$
\affiliations
$^1$University of New South Wales\\
$^2$University of Science and Technology of China\\
$^3$The Chinese University of Hong Kong\\
\emails
\{yu.hao, xin.cao, yixiang.fang\}@unsw.edu.au,
xkxie@ustc.edu.cn, 
swang@se.cuhk.edu.hk}

\begin{document}

\maketitle

\begin{abstract}
Predicting the link between two nodes is a fundamental problem for graph data analytics. In attributed graphs, both the structure and attribute information can be utilized for link prediction. Most existing studies focus on transductive link prediction where both nodes are already in the graph. However, many real-world applications require inductive prediction for new nodes having only attribute information. It is more challenging since the new nodes do not have structure information and cannot be seen during the model training. To solve this problem, we propose a model called \TriNN, which consists of three components: two node embedding encoders and one alignment mechanism. The two encoders aim to output the attribute-oriented node embedding and the structure-oriented node embedding, and the alignment mechanism aligns the two types of embeddings to build the connections between the attributes and links.
Our model \TriNN is versatile in the sense that it works for both inductive and transductive link prediction. Extensive experiments on several benchmark datasets show that our proposed model significantly outperforms existing inductive link prediction methods, and also outperforms the state-of-the-art methods on transductive link prediction.
\end{abstract}

\section{Introduction}
Link prediction is a fundamental task in graph data analytics~\cite{Liben-NowellK07}. Many real-world applications can benefit from link prediction, such as recommendations, knowledge graph completion, etc~\cite{bhagavatula2018content,kazemi2018simple,xu2019link}. Graph embedding, which represents the nodes on the graph by low-dimensional vectors, has been proved as an effective approach for link prediction on attributed graphs (e.g.,~\cite{kipf2017semi,bojchevski2018deep}).

\begin{figure}[!t]
    \centering
    \includegraphics[width=80mm]{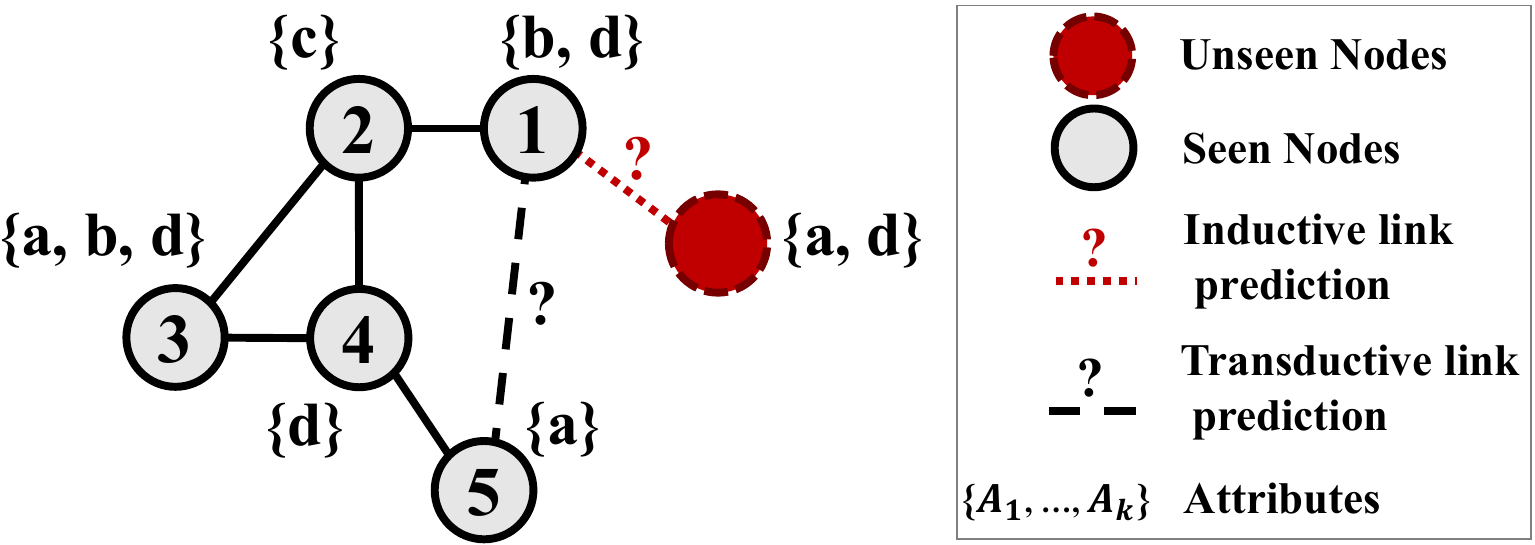}
    \caption{{\bf Inductive} and {\bf transductive} link predictions.}
    \label{fig:problem introduction}
\end{figure}

Generally, there are two types of link prediction, i.e., transductive and inductive, as shown in Figure~\ref{fig:problem introduction}. Most existing (attributed) graph embedding approaches focus on transductive link prediction, where both nodes are already in the given graph and can be seen during the training process, such as GCN~\cite{kipf2017semi}, GAE~\cite{kipf2016variational}, and SEAL~\cite{zhang2018link}. However, many real-world applications need inductive link prediction which requires embeddings to be quickly generated for new nodes with only attribute information (e.g., a new user in a recommender system). SDNE~\cite{wang2016structural} and GraphSage~\cite{hamilton2017inductive} can compute embeddings for new nodes but the edges of nodes are required. G2G~\cite{bojchevski2018deep} can perform inductive link prediction for the unseen nodes without local structures. However, it cannot well distinguish the nodes with similar attributes, because this model does not well capture the structure information in the node representations.

In this work, we propose a novel graph embedding model called \textit{\underline{D}ual-\underline{E}ncoder graph embedding with \underline{AL}ignment} (\TriNN) for inductive link prediction of new nodes with only attribute information. 
We aim to learn the connections between the nodes' attributes and the graph structure through this model. The model embeds the graph nodes into the vector space, and it can compute an embedding vector for the new query node with only attributes which is compared to another node's embedding for link prediction.

\begin{figure}[!h]
    \centering
    \includegraphics[width=7cm]{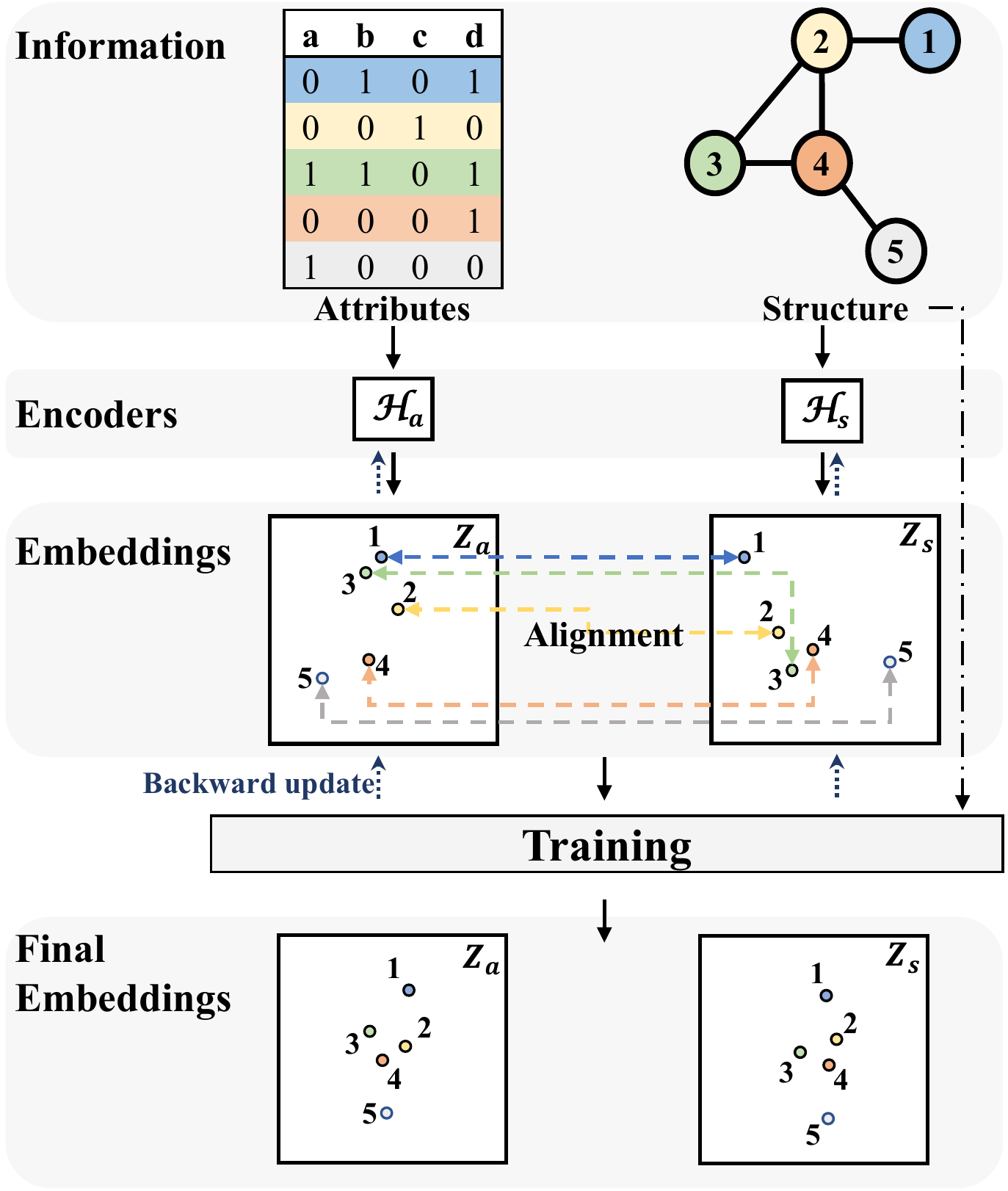}
    \caption{An illustration of the proposed model \TriNN.}
    \label{fig:framework overview}
\end{figure}

As shown in Figure~\ref{fig:framework overview}, \TriNN has three components: an attribute-oriented encoder $\mathcal{H}_a$, a structure-oriented encoder $\mathcal{H}_s$, and an alignment mechanism. The attribute-oriented encoder $\mathcal{H}_a$ maps the nodes' attributes to embeddings in the vector space. Given two linked attributed nodes, the similarity of their embedding vectors computed by $\mathcal{H}_a$ is high. $\mathcal{H}_a$ is used for computing the embedding vectors of the new nodes with attributes. $\mathcal{H}_s$ computes the node embeddings that preserve the structure information. Given two linked nodes, the similarity of their embedding vectors computed by $\mathcal{H}_s$ is high. The alignment mechanism aligns the two types of embeddings to build the connections between the attributes and graph structure. The two encoders keep being updated during the training process such that their embeddings produced are aligned. 
In addition, we use a novel ranking-motivated loss that can be regularized by hyper-parameters, which is not considered in existing ranking loss-based graph embedding models. Although \TriNN focuses on inductive link prediction with only attribute information, it can perform transductive link prediction using both structural and attribute information as well. 
The main contributions of our approach are summarized as follows:
\begin{itemize}
\item We design a model \TriNN for inductive link prediction of the new nodes with only attribute information given. 
\item The proposed {\bf alignment mechanism} builds the connections between attributes and graph structure, and improves the representation ability of node embeddings. 
\item The experimental results on several real-world datasets show that our proposed model consistently outperforms the state-of-the-art models on both inductive (with at least 6\% AP score improvement) as well as transductive link prediction.
\end{itemize}

\section{Related Works}
The early studies on link prediction usually have strong assumptions on when links may exist, and they are mainly based on different similarity measures such as the number of common neighbours, Jaccard coefficient, and Adamic-Adar to measure the node proximity
~\cite{Liben-NowellK07,zhou2009predicting,ZhaoAH16}. They assume that the probability of the existence of the link between two nodes increases with their similarity. However, such assumptions may not hold in some real-world networks such as the protein-protein interaction (PPI) networks, because two proteins sharing many common neighbours are actually less likely to interact~\cite{kovacs2019network}.

Recently, researchers have shown an increasing interest in solving the link prediction problem via graph embedding method~\cite{zhang2018link}. Graph embedding has been used widely~\cite{gao2018deep,gao2019progan,qiutemporal,8970926}, and it aims to map nodes $\mathcal{V}$ to $l$-dimensional vectors $\mathcal{Z}=\{\mathbf{z}^1,...,\mathbf{z}^n\}, \mathbf{z}^i \in \mathbb{R}^{l}$.
Some graph embedding methods only capture the structural information of the graph, such as the random walk-based graph embedding approaches DeepWalk~\cite{perozzi2014deepwalk} and node2vec~\cite{grover2016node2vec} that adapt the Skip-Gram model and treat each generated path of nodes as the sequence of words, and SDNE~\cite{wang2016structural} that learns node embedding preserving both local and global structures. They do not leverage the attribute information and thus cannot be used for link prediction in attributed graphs.

For link prediction in attributed graphs, most existing studies focus on transductive link prediction, where both nodes are already in the graph. For example, GCN~\cite{kipf2017semi} requires that the full graph Laplacian is known during training; GAE~\cite{kipf2016variational} learns the node embeddings with a GCN encoder and an inner product decoder. SEAL~\cite{zhang2018link} is another GCN-based framework that solves the link prediction problem using local sub-graphs.
To our best knowledge, only G2G~\cite{bojchevski2018deep} is able to perform the task of inductive link prediction with only attribute information. It relies on a deep encoder that embeds each node as a Gaussian distribution. However, the output of its encoder is primarily based on the nodes' attributes, and thus it cannot well distinguish the nodes with similar attributes. G2G will embed these nodes nearby in the vector space, because their structure information is not fully utilized.

In our model \TriNN, the alignment mechanism aligns the attribute embedding and the structure embedding, and both attribute and structure information can be well captured by the learned node representations, and thus it can achieve better link prediction performance.

\section{Problem Statement}
We study the problem on an attributed graph $G = (\mathcal{V},\mathcal{E},\mathcal{X})$, with a node set $\mathcal{V} = \{ v_1, ... ,v_n \}$ ($n=|\mathcal{V}|$), an edge set $\mathcal{E} \subseteq \mathcal{V} \times \mathcal{V}$, and an attribute feature matrix $\mathcal{X} = \{\mathbf{x}_1, ... ,\mathbf{x}_n\}$, where $\mathbf{x}_i \in \mathbb{R}^m$ is the attribute feature vector of node $v_i$ ($m$ is the number of attributes in the graph). 

Given a graph $G$ as well as a pair of nodes $(v_p,v_q)$, the {\bf link prediction} task aims to predict the existence of the link between $v_p$ and $v_q$. 
In {\bf transductive} link prediction, both nodes $v_p$ and $v_q$ are already in the graph $G$ (i.e., $v_p, v_q \in \mathcal{V}$ and thus they can be seen during the training process). In this work, we focus on the {\bf inductive} link prediction, where either or both $v_p$ and $v_q$ are not seen during the training process. During the prediction, only the attribute information of the new nodes is available, i.e., their local structures are unknown.

We propose to utilize the node embedding approach, which embeds the graph nodes into the vector space, to solve this problem.  When performing prediction, for a new node $v_q$ with its attribute feature vector $\mathbf{x}_q\in \mathbb{R}^m$, the model outputs the node embedding for this node $v_q$ from $\mathbf{x}_q$ and then compares $v_q$'s embedding with another node's embedding to predict their relationship.

\section{Our Model \TriNN}\label{sec:framework}
\subsection{Overview}
Figure~\ref{fig:framework overview} illustrates our model \TriNN, which consists of three components: two node embedding encoders and one alignment mechanism. We aim to predict the links for new nodes with only attribute information. This requires us to build the connections between the attributes and the links between nodes, which means that we need to compute an embedding vector for a given set of attributes. Our attribute-oriented encoder $\mathcal{H}_a$ performs this task. It maps the nodes' attributes to embeddings in the vector space. Given two linked nodes with their attributes, the similarity of their embedding vectors computed by $\mathcal{H}_a$ is high.

However, when the nodes attributes are too uninformative or similar, such a single encoder is not sufficient to output useful embeddings to distinguish nodes with similar attributes, because they are all embedded nearby in the vector space. To remedy this issue, we use another structure-oriented encoder $\mathcal{H}_s$ to compute node embeddings that preserve the graph structure information. As a result, given two linked nodes, the similarity of their embedding vectors computed by $\mathcal{H}_s$ is high.
Next, we propose an alignment mechanism to align the two types of embeddings. During the training process, the two encoders keep being updated in order to produce node embeddings that are aligned in the vector space. Finally, the connections between the attributes and the links are captured by the two encoders, yielding better embedding vector computation during the link prediction. 

\subsection{Attribute-oriented Encoder}
The attribute-oriented encoder $\mathcal{H}_a$ takes a node attribute vector $\mathbf{x}_i$ (attributes of the node $v_i$) as the input and outputs a node embedding, denoted as $\mathbf{z}_a^i$, i.e., $\mathbf{z}_a^i = \mathcal{H}_a(\mathbf{x}_i)$.

There are many neural network choices for learning $\mathcal{H}_a$, and we choose the multilayer perceptron (MLP) with nonlinear activation layer  ~\cite{Goodfellow-et-al-2016} in this paper as follows:
\begin{align}
    \mathcal{H}_a(\mathbf{x}_i) = \sigma(\mathbf{W}_a^2(\sigma(\mathbf{W}_a^1\mathbf{x}_i+\mathbf{b}_a^1))+\mathbf{b}_a^2), 
\label{eq:H_a}
\end{align}
with trainable parameters $\mathbf{W}_a^1$, $\mathbf{W}_a^2$, $\mathbf{b}_a^1$ and $\mathbf{b}_a^2$, and the exponential linear unit $\sigma(\cdot)$~\cite{clevert2015fast}. It is to be noted that the layer number of $\mathcal{H}_a$ varies with different datasets.
Here, we do not use GCN in $\mathcal{H}_a$, because we observe that aggregating too much information from a node's neighbours may affect the representation ability of the node's attributes for the link prediction task in attributed graphs. As shown in Section~\ref{sec:transductive result}, we tried GCN-like encoders to obtain node embeddings from attributes, and the performance is not as good as using MLP. In addition, without aggregating the information from neighbours, the training process can be speeded up significantly.

\subsection{Structure-oriented Encoder}
The structure-oriented encoder $\mathcal{H}_s$ aims to generate node embeddings that preserve the structural information of the graph without considering the attributes. We expect that the vectors of two linked nodes computed by $\mathcal{H}_s$ can have high similarity.
To achieve this, we use the one-hot encoding of the nodes $\mathcal{I_V}=\{\mathbf{I}_1,...,\mathbf{I}_n\}$ (can be regarded as nodes' identifiers~\cite{you2019position}) as the input of $\mathcal{H}_s$. $\mathcal{H}_s$ can be seen as a function that maps node $v_i$ to its node embedding vector $\mathbf{z}_s^i$: $\mathbf{z}_s^i = \mathcal{H}_s(\mathbf{I}_i)$.
From the perspective of Occam’s razor, we adopt a linear model as the encoder $\mathcal{H}_s$:
\begin{equation}
    \mathcal{H}_s(\mathbf{I}_i) = g(\mathbf{W}_s)\: \mathbf{I}_i, 
\end{equation}
where the weight normalization~\cite{salimans2016weight} $g(\cdot)$ is employed to reparameterize the parameter $\mathbf{W}_s$. In addition, $g(\cdot)$ is able to accelerate the convergence of stochastic gradient descent optimization. By minimizing the ranking-motivated loss which will be presented in details in Section~\ref{sec:training}, the final embedding $\mathcal{Z}_s$ is able to capture the structural information.

Note that the encoder $\mathcal{H}_s$ can be another node embedding method in \TriNN that focuses on learning graph structure with corresponding input, such as GCN with the adjacency matrix as the input. However, we find that using GCN in $\mathcal{H}_s$ decreases the link prediction performance, which is shown in Tables~\ref{tab:inductive-result} and~\ref{tab:transductive-result} in Section~\ref{sec:experiments}. 

\subsection{Alignment Mechanism and Model Training} \label{sec:training}

We propose an alignment mechanism to align the embeddings generated by the two types of encoders to learn the connections between the node attributes and the graph structure. During the model training, the two encoders keep being updated in order to be able to produce the embeddings that are aligned in the vector space. We first introduce the loss functions we used, and then we describe the proposed alignment mechanism in our model, and we finally present the training algorithm for \TriNN.

\paragraph{Loss Function.}
Learning graph embedding via ranking is based on the ranking-motivated loss, which has been proved to be effective in many studies \cite{bojchevski2018deep,bhagavatula2018content}. In this paper, we propose a novel mini-batch learning method with a personalized ranking-motivated loss to learn the node embeddings with comprehensive representation performance.

Optimizing a ranking-motivated loss can help to capture the relationships between each pair of training samples. Contrastive loss \cite{hadsell2006dimensionality} is one kind of ranking-motivated loss, and it is originally proposed to solve the dimensionality reduction problem. Given $k$ pairs of samples $\{(p_1,q_1),..., (p_k, q_k) \}$, the loss is shown as follows:
\begin{align}
    \mathcal{L}_{c} =  \frac{1}{2k}\sum_{i=1}^{k} &(1-y_i) \, \max(0,\tau -d(p_i,q_i))^2
    \nonumber\\
    & +  y_i\,d(p_i,q_i)^2,
\label{eq:contrastive}
\end{align}
where $d(\cdot,\cdot)$ is a distance function, $y_i=1$ if the samples $p_i$ and $q_i$ are deemed similar and $y_i=0$ otherwise, $\max(0,\cdot)$ is the hinge function, and $\tau$ is the margin. The learning objective of Eq.~\ref{eq:contrastive} is to map similar input samples to nearby points in the output vector space and map dissimilar samples to distant points. 

We have a similar learning objective in our problem that is to map the linked nodes (positive sample) in the graph to points that are close in the output vector space and map the unlinked nodes (negative sample) to points that are far away from each other.
However, there is a problem when directly adapting contrastive loss for link prediction. The negative pair-wise samples have different distances in the graph, and thus using a fixed margin for all the negative samples in Eq.~\ref{eq:contrastive} is not appropriate, but it is difficult to set proper margins for the negative samples with different distances. 
Moreover, neither the existing work \cite{bojchevski2018deep,bhagavatula2018content} nor Eq.~\ref{eq:contrastive} considers the regularization in the loss function, which is important and can further improve the prediction performance. Motivated by the above observations, we propose the following loss to be optimized for a given mini-batch of node pair samples $B = \{(v_{p_1},v_{q_1}),..., (v_{p_k},v_{q_k})\}$ where $p_i\neq q_i$ with $i \in [1,k]$
(obtained by sampling $k$ pairs of nodes from the graph):
\begin{align}
    \mathcal{L}_B(\mathcal{Z}) =  \frac{1}{\left |B  \right |}\sum_{(v_{p_i},v_{q_i}) \in B}&
    \left [ \right .(1-y_i)\alpha(v_{p_i},v_{q_i}) \phi_1(-s(\mathbf{z}^{p_i},\mathbf{z}^{q_i}))
    \nonumber\\
    + & y_i\phi_2(s(\mathbf{z}^{p_i},\mathbf{z}^{q_i}))\left .\right ],
\label{eq:single_loss}
\end{align}
where $s(\cdot,\cdot)$ is the function to measure the similarity between node embeddings $(\mathbf{z}^{p_i}$ and $\mathbf{z}^{q_i})$, $y_i$ is the link relation label with $y_i = 1$ if $v_{p_i}$ and $v_{q_i}$ are connected and $y_i = 0$ otherwise, $\alpha(\cdot,\cdot)$ is a weight function, and $\phi_1(\cdot)$ and $\phi_2(\cdot)$ are derived from the function $\phi(\cdot)$ with different hyper-parameters. Specifically, there are many choices of $s(\cdot,\cdot)$, such as dot product, cosine similarity, etc. A high score of $s(\cdot,\cdot)$ indicates that the nodes are similar, and vice versa. We find that using the cosine similarity have good results in our model. %As for the mini-batch $B$, we randomly sample $k$ pairs of nodes over the graph $G$, and the ratio $r$ of linked node pairs in each mini-batch is parameterized. 

We use $\phi_1(\cdot)$ and $\phi_2(\cdot)$ because the regularization is not considered in Eq.~\ref{eq:contrastive}. 
Inspired by that the logistic loss can be seen as a ``soft'' version of the hinge loss with an infinite margin \cite{lecun2006tutorial} and is differentiable, we adopt the generalized logisitic loss function $\phi(x)$ as follows:
\begin{equation}
\phi(x) = \frac{1}{\gamma}\log(1+e^{-\gamma x+b}),
\label{eq:log-loss}
\end{equation}
where $\gamma>0$ and $b$ are loss margin parameters that can tune regularization \cite{masnadi2015view}. 

$\alpha(\cdot,\cdot)$ is a weight function to measure the importance of negative samples with different distances. Specifically, we define $\alpha(\cdot,\cdot)$ as follows:
\begin{equation}
    \alpha(v_{p},v_{q}) = \exp(\frac{\beta}{d_{sp}(v_{p},v_{q})}),
    \label{eq:alpha}
\end{equation}
where $\beta > 0$ is a hyper-parameter, and $d_{sp}(\cdot,\cdot)$ denotes the shortest path distance between a node pair. If node $v_p$ cannot reach node $v_q$, $d_{sp}(v_p,v_q) = \infty$. This weight aims to help the model pay more attention to the close negative neighbours during the training process.

\paragraph{Alignment Mechanism.}\label{sec:alignment}
By minimizing $\mathcal{L}_B(\mathcal{Z}_s)$ and $\mathcal{L}_B(\mathcal{Z}_a)$, we can lean structure-oriented node embedding $\mathcal{Z}_s$ and attribute-oriented node embedding $\mathcal{Z}_a$, respectively. However, if we learn them separately, the two types of embeddings are isolated and cannot well represent the connections between attributes and graph structure. We propose to align the two types of embeddings during the training process and learn the two encoders simultaneously. We design two alignment methods:

\noindent1. \underline{Tight Alignment} ($T$-$align$) aims to maximize the similarity between $\mathbf{z}_s^{i}$ and $\mathbf{z}_a^{i}$ for each node $v_i$. Mathematically, the objective of the tight alignment is to minimize
\begin{equation}
    \mathcal{L}_{T\text{-}align}(\mathcal{Z}_s,\mathcal{Z}_a) = -\frac{1}{\left | V \right |}\sum_{v_i\in V}s(\mathbf{z}_s^{i},\mathbf{z}_a^{i}).
\end{equation}
However, the tight method sometimes is too strict during the aligning the two types of embeddings. 

\noindent2. \underline{Loose Alignment} ($L$-$align$) aims to maximize the similarity between $\mathbf{z}_s^{p_i}$ and $\mathbf{z}_a^{q_i}$ of two linked nodes $v_{p_i}$ and $v_{q_i}$, and it adopts the loss function in Eq.~\ref{eq:single_loss}. Mathematically, the objective of the loose alignment is to minimize
\begin{align}
    \mathcal{L}_{L\text{-}align}&(\mathcal{Z}_s,\mathcal{Z}_a) = \frac{1}{\left |B  \right |}\sum_{(v_{p_i},v_{q_i}) \in B}
    \left [ \right . y_i\phi_2(s(\mathbf{z}_s^{p_i},\mathbf{z}_a^{q_i}))
    \nonumber\\
    + & (1-y_i)\alpha(v_{p_i},v_{q_i})\phi_1(-s(\mathbf{z}_s^{p_i},\mathbf{z}_a^{q_i})) \left .\right ]
\end{align}

Putting everything all together, the final objective of our model is as below:
\begin{equation}
    \mathcal{L}= \theta_1\mathcal{L}_B(\mathcal{Z}_s)+\theta_2\mathcal{L}_B(\mathcal{Z}_a)+\theta_3\mathcal{L}_{align}(\mathcal{Z}_s,\mathcal{Z}_a),
\label{eq:final_loss}
\end{equation}
where $\boldsymbol\theta = [\theta_1,\theta_2,\theta_3]$ is a hyper-parameter vector to parameterize the weights of different losses.

\paragraph{Training algorithm and prediction.} Algorithm~\ref{alg:algorithm} summarizes the training process of the proposed model.

\setlength{\textfloatsep}{1pt}
\begin{algorithm}[!t]
\caption{The learning process of \TriNN}
\label{alg:algorithm}
\textbf{Input}: Graph $G = (\mathcal{V},\mathcal{E},\mathcal{X})$; a set of mini-batches $\mathcal B$;
loss weight $\boldsymbol\theta$ and other hyper-parameters;
\\
\textbf{Output}: Node embeddings $\mathcal{Z}_s$ and $\mathcal{Z}_a$

\begin{algorithmic}[1] %[1] enables line numbers
\FOR {$B \in \mathcal B $}
\STATE $\mathcal{Z}_s \leftarrow  \mathcal{H}_s(\mathcal{I_V})$
\STATE $\mathcal{Z}_a \leftarrow  \mathcal{H}_a(\mathcal{X})$
\STATE $\mathcal{L} \leftarrow \boldsymbol\theta \cdot \left [ \mathcal{L}_B(\mathcal{Z}_s), \mathcal{L}_B(\mathcal{Z}_a), \mathcal{L}_{align}(\mathcal{Z}_s,\mathcal{Z}_a) \right ]$
\STATE Update $\mathcal{H}_s$ and $\mathcal{H}_a$ with  stochastic gradient $\bigtriangledown \mathcal{L}$
\ENDFOR
\STATE Update $\mathcal{Z}_s$ and $\mathcal{Z}_a$ via $\mathcal{H}_s$ and $\mathcal{H}_s$
\STATE \textbf{return} $\mathcal{Z}_s$ and $\mathcal{Z}_a$
\end{algorithmic}
\end{algorithm}
To predict whether there is a link between two nodes $v_p$ and $v_q$, we can calculate a score with $\mathcal{Z}_s$ and $\mathcal{Z}_a$ as follows
\begin{equation}
\resizebox{.89\linewidth}{!}{$
    \displaystyle
        score(v_p,v_q) = \lambda_1 s(\mathbf{z}_s^p,\mathbf{z}_s^q)
                +\lambda_2 s(\mathbf{z}_a^p,\mathbf{z}_a^q)
                +\lambda_3 s(\mathbf{z}_s^p,\mathbf{z}_a^q)
$,}
\label{eq:predict}
\end{equation}
where $\boldsymbol\lambda = [\lambda_1,\lambda_2,\lambda_3]$ is another hyper-parameter vector used to give each similarity score a different weight. In inductive link prediction, for a new node $v_q$, $\mathbf{z}_a^q$ is computed by $\mathcal{H}_a$, and $\lambda_1=0$. Our model can also perform transductive link prediction by setting $\lambda_1$ to a non-zero value.

\section{Experiments}\label{sec:experiments}
\subsection{Datasets}

\begin{table}[!h]
\centering
\scalebox{0.88}{
\small
\begin{tabular}{lrrr}
\toprule
\textbf{Datasets} & \textbf{Nodes} & \textbf{Edges} & \textbf{Attributes} \\
\midrule
CS (\cite{shchur2018pitfalls})& 18,333 & 81,894 & 6,805 \\
PPI (\cite{Zitnik2017})& 1,767 & 16,159 & 50 \\
Cora (\cite{mccallum2000automating})& 2,708 & 5,278 & 1,433 \\
CiteSeer (\cite{sen2008collective})& 3,327 & 4,552 & 3,703 \\
PubMed (\cite{namata2012query})& 19,717 & 44,324 & 500 \\
Computers (\cite{mcauley2015image})& 13,752 & 245,861 & 767 \\
Photo (\cite{mcauley2015image})& 7,650 & 119,081 & 745 \\
\bottomrule
\end{tabular}
}
\caption{Statistics of experimental datasets.}
\label{tab:datasets}
\end{table}

\noindent For link prediction tasks, we evaluate our proposed model and baselines on four types of real-world datasets, i.e., the co-authorship graph (\textbf{CS}), the protein-protein interactions graph (\textbf{PPI}), co-purchase graphs (\textbf{Computers} and \textbf{Photo}), and citation network datasets (\textbf{Cora}, \textbf{CiteSeer} and \textbf{PubMed}). Details of these datasets are summarised in Table~\ref{tab:datasets}.

\begin{table*}[!t]
% \small
\centering
\scalebox{1}{
\begin{tabular}{lcccccccccccc}
\toprule
 & \multicolumn{2}{c}{Cora} & \multicolumn{2}{c}{CiteSeer} & \multicolumn{2}{c}{CS} & \multicolumn{2}{c}{PubMed} & \multicolumn{2}{c}{Computers} & \multicolumn{2}{c}{Photo} \\
 & AUC & AP & AUC & AP & AUC & AP & AUC & AP & AUC & AP & AUC & AP \\
\midrule
MLP & 0.826 & 0.674 & 0.897 & 0.789 & 0.921 & 0.810 & 0.842 & 0.705 & 0.866 & 0.692 & 0.901 & 0.753 \\
Cite. & 0.839 & 0.712 & 0.914 & 0.824 & 0.939 & 0.862 & 0.912 & 0.809 & 0.898 & 0.762 & 0.926 & 0.808 \\
G2G & 0.845 & 0.739 & 0.922 & 0.842 & 0.948 & 0.889 & 0.910 & 0.798 & 0.853 & 0.684 & 0.862  & 0.704  \\
\midrule
% \hline
GCN-\TriNN & 0.855 & 0.766 & 0.912 & 0.862 & 0.969 & 0.943 & 0.961 & 0.924 & 0.943 & 0.888 & 0.959  & 0.907  \\
\TriNN & \textbf{0.864} & \textbf{0.804} & \textbf{0.937} & \textbf{0.907} & \textbf{0.977} & \textbf{0.959} & \textbf{0.966} & \textbf{0.931} & \textbf{0.953} & \textbf{0.899} & \textbf{0.965} & \textbf{0.922}\\
\bottomrule
\end{tabular}
}
\caption{The results of inductive link prediction.}
\label{tab:inductive-result}
\end{table*}

\subsection{Baseline Methods}
We compare our model \textbf{\TriNN} with \textbf{MLP} and several state-of-the-art graph embedding methods, including \textbf{SEAL}~\cite{zhang2018link}, \textbf{G2G}~\cite{bojchevski2018deep} and \textbf{GAE}~\cite{kipf2016variational}. In addition, the original GAE takes {\bf GCN} as the encoder. We also consider other GAE variants, which replace the GCN encoder with {\bf GIN} \cite{xu2018how}, {\bf GAT} \cite{velickovic2018graph} and {\bf SAGE} \cite{hamilton2017inductive} respectively. The GAE variants are denoted as their encoder model names.

Moreover, \TriNN variants can use different graph embedding models as structure-oriented or attributed-oriented encoders. \TriNN denotes the proposed encoders presented in Section~\ref{sec:framework}. A \TriNN variant is denoted as $X$-$Y$, using the $X$ model as the structure-oriented and the $Y$ model as the attribute-oriented encoder. We select three representative variants. All the \TriNN variants aim to minimize Eq.~\ref{eq:final_loss}. To ensure fairness, we set all models with a similar amount of parameters and train them for the same number of epochs.

\subsection{Experimental Setup}
We evaluate the proposed model \TriNN and baseline models under both inductive and transductive learning settings. 

 \paragraph{Inductive link prediction.} For the inductive case, the nodes in the test set are unseen during the training process. Similar to the dataset split setting of \cite{bojchevski2018deep}, we randomly hide 10\% nodes and use the edges between them for the test set. The remaining nodes and edges are used for training and validation.   

\paragraph{Transductive link prediction.} For the transductive case, all the nodes on the graph can be seen during the training. Similar to the dataset split setting of \cite{you2019position}, we randomly sample 10\%/10\% edges and an equal number of non-edges as validation/test set. The remaining non-edges and 80\% edges are used as the training set. 

The test set performance will be reported when the model achieves the best performance on the validation set. For the experimental results, we report the mean area under the ROC curve (AUC) and the average precision (AP) scores over ten trials with different random seeds and train/validation splits.
In all the experiments, the default embedding size is 64. For each training mini-batch, the linked node pairs account for 40\%. We tune the hyper-parameters of baseline models and our proposed \TriNN with the grid search algorithm on the validation set.

\subsection{Results of Inductive Link Prediction}\label{sec:inductive result}
 As GCN-based models cannot aggregate neighbours' information in the inductive link prediction scenario, we compare our proposed model with MLP and G2G, and the experimental results are shown in Table~\ref{tab:inductive-result}. It shows that \TriNN significantly outperforms MLP and G2G across all datasets. On the Computers dataset, for instance, \TriNN improves AUC and AP scores by 6.12\% and 17.98\%, respectively. By comparing GCN-\TriNN with \TriNN, it shows that using GCN-layer as $\mathcal{H}_s$ cannot improve the link prediction performance. Also, it is observed that G2G performs worse when the graphs contain a small number of feature dimensions (compared with the number of nodes), such as Computers and Photo. The reason is that the node embedding encoder of G2G solely takes feature matrix $\mathbf{X}$ as the input. As an extreme example, when the features of all the nodes are similar, it will be difficult to distinguish different nodes for G2G.

\begin{table*}[!t]
\centering
\scalebox{1}{
\begin{tabular}{lcccccccccc}
\toprule
 & \multicolumn{2}{c}{Cora} & \multicolumn{2}{c}{CiteSeer} & \multicolumn{2}{c}{CS} & \multicolumn{2}{c}{PubMed} & \multicolumn{2}{c}{PPI} \\
 & AUC & AP & AUC & AP & AUC & AP & AUC & AP & AUC & AP \\
\midrule
GAT & 0.8684 & 0.8866 & 0.8423 & 0.8662 & 0.9465 & 0.9473 & 0.9193 & 0.9202 & 0.8092 & 0.8136 \\
GCN & 0.8670 & 0.8755 & 0.8466 & 0.8620 & 0.9452 & 0.9421 & 0.9287 & 0.9272 & 0.8384 & 0.8364 \\
GIN & 0.8666 & 0.8762 & 0.8405 & 0.8617 & 0.9432 & 0.9407 & 0.9262 & 0.9254 & 0.8086 & 0.8086 \\
SAGE & 0.8739 & 0.8881 & 0.8498 & 0.8721 & 0.9485 & 0.9504 & 0.9254  & 0.9270 & 0.8112 & 0.8131 \\
Cite. & 0.9145 & 0.9143 & 0.9385 & 0.9417 & 0.9501  & 0.9517 & 0.9435 & 0.9378 & 0.6047  & 0.5981\\
SEAL & 0.8269 & 0.7959 & 0.8064 & 0.7769 & 0.9146 & 0.8856 & 0.9235 & 0.9239 & 0.8825 & 0.8749 \\
P-GNN & 0.8225 & 0.8427 & 0.8065 & 0.8436 & 0.8779 & 0.8811 & 0.8145 & 0.8647 & 0.7303 & 0.6716\\
G2G & 0.9282 & 0.9336 & 0.9413 & 0.9421 & 0.9636 & 0.9640 & 0.9432 & 0.9364 & 0.5896 & 0.5333 \\
\midrule
% \hline
GCN-\TriNN  & 0.9163 & 0.9047 & 0.9221 & 0.9219 & 0.9796 & 0.9801 & 0.9456 & 0.9498 & 0.8673 & 0.8709 \\
\TriNN-GCN & 0.9002 & 0.9075 & 0.8496 & 0.8747 & 0.9646 & 0.9663 & 0.9299 & 0.9348 & 0.8861 & 0.8868 \\
\TriNN-GAT & 0.8985 & 0.9091 & 0.8463 & 0.8752 & 0.9640 & 0.9666 & 0.9311 & 0.9355 & 0.8711 & 0.8762 \\
\TriNN & \textbf{0.9455} & \textbf{0.9501} & \textbf{0.9519} & \textbf{0.9591} & \textbf{0.9827} & \textbf{0.9841} & \textbf{0.9593} & \textbf{0.9611} & \textbf{0.8894} & \textbf{0.8973} \\
\bottomrule
\end{tabular}
}
\caption{The results of transductive link prediction.}
\label{tab:transductive-result}
\end{table*}

\subsection{Results of Transductive Link Prediction}\label{sec:transductive result}
The experimental results of transductive link prediction are summarized in Table~\ref{tab:transductive-result}. The results show that our proposed model \TriNN achieves the best performance. For the baselines, G2G performs well on the citation networks and co-authorship graph, which have informative node attributes. It is worth noting that the number of node attributes in PPI is less than 10\% of that in other datasets. In the PPI graph, where each node contains limited attribute information, G2G has the worst performance, while SEAL achieves outstanding performance.

Interestingly, the remaining GAE baseline models achieve comparable performance on all the datasets, although they have different methods of aggregating neighbours' information. Moreover, the attention mechanism of GAT does not outperform other GNN layers in this scenario. The reason may be that the GAE variants are insensitive to the different information aggregation methods on the link prediction problem. Compared to the baselines, \TriNN is more robust and shows stronger generalization ability on different types of datasets. In addition, the \TriNN framework enables GAEs to achieve better performance.

\subsection{Comparison of Alignment Methods}
To compare different alignment methods in Section~\ref{sec:alignment}, we conduct both inductive and transductive link prediction experiments on three representative datasets, i.e., Cora, CS, and PubMed. The experimental results 
(Table~\ref{tab:alignment})  show that, on these three datasets, both two alignment methods are effective, and the loose alignment method slightly outperforms the tight alignment method, especially for the inductive link prediction task. The reason is that the loose alignment method places fewer restrictions on the node embedding alignment. 
The loose one also provides flexibility that the node embeddings $\mathcal{Z}_a$ and $\mathcal{Z}_s$ need by adjusting the hyper-parameters.

\begin{table}[tb]
\small
\centering
\scalebox{0.99}{
\begin{tabular*}{\hsize}{@{}@{\extracolsep{\fill}}ccccccc@{}}
\toprule
 & \multicolumn{2}{c}{Cora} & \multicolumn{2}{c}{CS} & \multicolumn{2}{c}{PubMed} \\
 & AUC & AP & AUC & AP & AUC & AP \\
\midrule
$T\text{-}align$-I & 0.845 & 0.774 & 0.972 & 0.951 & 0.951 & 0.905 \\
$L\text{-}align$-I & \textbf{0.865} & \textbf{0.803} & \textbf{0.976} & \textbf{0.956} & \textbf{0.966} & \textbf{0.931}\\
\midrule
% \hline
$T\text{-}align$-T & 0.939 & 0.942 & 0.976 & 0.978 & \textbf{0.955} & 0.958 \\
$L\text{-}align$-T & \textbf{0.946} & \textbf{0.950} & \textbf{0.983} & \textbf{0.984} & 0.954 & \textbf{0.961} \\
\bottomrule
\end{tabular*}
}
\caption{Comparison of different alignment methods. We consider both inductive and  transductive link prediction tasks, which are denoted as $align$-I and $align$-T respectively.}
\label{tab:alignment}
\end{table}

\begin{table}[tb] 
\small
\centering
\scalebox{0.99}{
\begin{tabular*}{\hsize}{@{}@{\extracolsep{\fill}}ccccccc@{}}
\toprule
 & \multicolumn{2}{c}{Cora} & \multicolumn{2}{c}{CS} & \multicolumn{2}{c}{PPI} \\
 $(\gamma,\alpha)$ & AUC & AP & AUC & AP & AUC & AP \\
\midrule
$(1,1)$ & 0.932 & 0.933 & 0.952 & 0.950 & 0.813 & 0.820 \\
$(1,\alpha^*)$ & 0.934 & 0.941 & 0.962 & 0.960 & 0.848 & 0.851 \\
$(\gamma^*,1)$ & 0.939 & 0.943 & 0.973 & 0.978 & 0.869 & 0.875 \\
$(\gamma^*,\alpha^*)$ & \textbf{0.946} & \textbf{0.950} & \textbf{0.983} & \textbf{0.984} & \textbf{0.889} & \textbf{0.897}\\
\bottomrule
\end{tabular*}
% \end{tabular}
}
\caption{The results of transductive link prediction with different values of $\gamma$ and $\alpha$. Here, $\gamma^*$ and $\alpha^*$ denote the optimal hyper-parameter values found by grid search algorithm on each dataset.}
\label{tab:loss}
\end{table}

\subsection{Parameter Analysis}
\begin{figure}[tb]
\begin{minipage}[b]{\linewidth}
\centering
\includegraphics[width=0.85\linewidth]{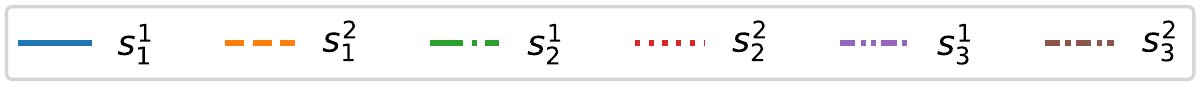}%
\end{minipage}
\centering
\subfigure[$\mathcal{Z}_s$]{
\begin{minipage}[t]{0.5\linewidth}
\centering
\includegraphics[width=38mm]{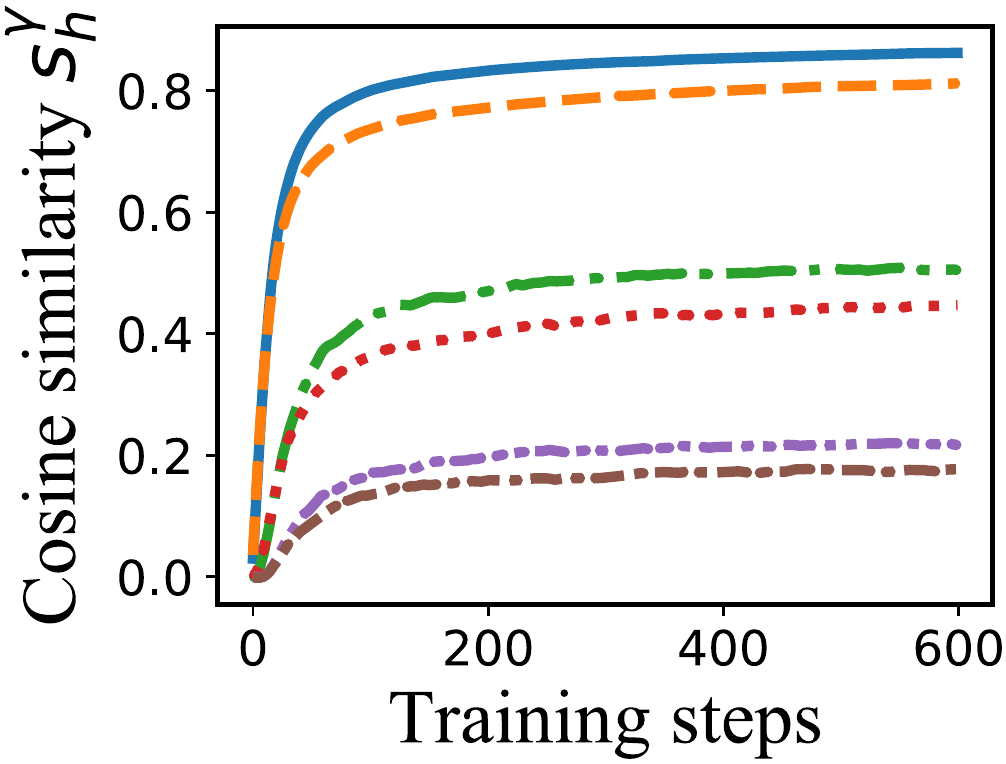}
\end{minipage}%
}%
\subfigure[$\mathcal{Z}_a$]{
\begin{minipage}[t]{0.5\linewidth}
\centering
\includegraphics[width=38mm]{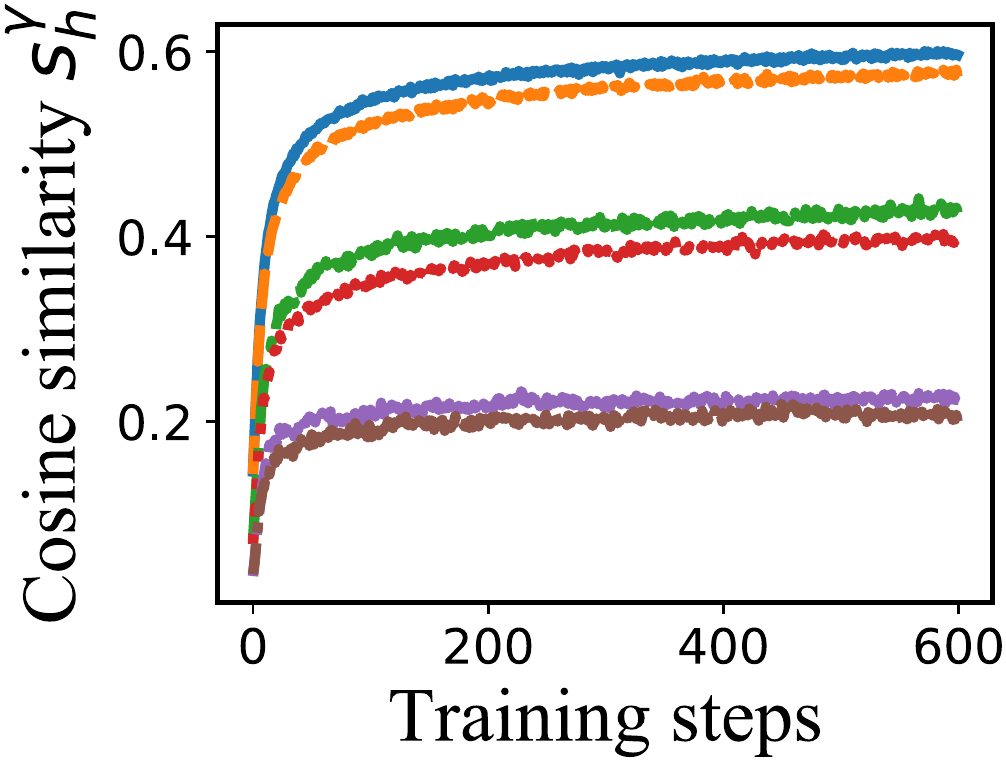}
\end{minipage}%
}%
\centering
\caption{Comparison of similarity scores with different values of $\gamma$. Given a set of $h$-hop node pairs $B_{h} = \{ (v_{p_i},v_{q_i})|d_{sp}(v_{p_i},v_{q_i})=h\}$, node embedding $\mathcal{Z}$ and $\gamma$, $s_h^\gamma = \frac{1}{|B_h|} \sum_{(v_{p_i},v_{q_i})\in B_h}cos(\mathbf{z}^{p_i},\mathbf{z}^{q_i})$ denotes the average cosine similarity of the pairs in $B_h$. Other hyper-parameters in \TriNN are fixed. Here, we compare $s_h^\gamma$ with $\mathcal{Z}_s$ and $\mathcal{Z}_a$ respectively for transductive link prediction task on CiteSeer.\\}
\label{fig:emb_margin}
\end{figure}

We here conduct experiments to analyse two key parameters, $\gamma$ (Eq.~\ref{eq:log-loss}) and $\alpha(\cdot,\cdot)$ (Eq.~\ref{eq:alpha}), in \TriNN. 
 Table~\ref{tab:loss} indicates that the performance can be improved with tuning both of them, and $\gamma$ plays a more important role than $\alpha$. The reason is that varying $\gamma$ is able to regularize the loss (Eq.~\ref{eq:single_loss}). It is interesting to note that different $\gamma$ can also change the similarity of node pairs in the embedding space, as shown in Figure~\ref{fig:emb_margin}. It also indicates that the same node pair tends to have a higher similarity score in the structure-oriented embedding space than the one in the attribute-oriented embedding space. The reason is that the node embeddings in $\mathcal{Z}_s$ are separate, while there are correlations between the ones in $\mathcal{Z}_a$, especially for those who have certain common attributes.

\section{Conclusions}
In this work, we propose a novel model \TriNN to address the inductive link prediction problem on attributed graphs, where the local structure of the new node is unknown. Different from the typical GCNs that aggregate information from neighbours, \TriNN learns comprehensive node representations via two encoders and an alignment mechanism. We have experimentally shown that our proposed model \TriNN consistently outperforms state-of-the-art methods. In the future, we will develop more efficient training algorithms, so that our model can process large-scale datasets.

\section*{Acknowledgments}
Xin Cao is supported by ARC DE190100663. Xike Xie is supported by NSFC (No. 61772492), Jiangsu NSF (No. BK20171240) and the CAS Pioneer Hundred Talents Program. Sibo Wang is supported by Hong Kong RGC ECS Grant (No. 24203419), CUHK Direct Grant (No. 4055114), and NSFC (No. U1936205). 
% \newpage
\bibliographystyle{named}
% \small
\bibliography{ijcai20}

\end{document}